\documentclass[letterpaper, 10 pt, conference]{ieeeconf/ieeeconf}
\IEEEoverridecommandlockouts
\usepackage{cite}
\usepackage{amsmath,amssymb,amsfonts}
\usepackage{algorithmic}
\usepackage{graphicx}
\usepackage{textcomp}
\usepackage{xcolor}
\usepackage{hyperref}
\usepackage{booktabs}
\usepackage{multirow}
\usepackage{subcaption}
\usepackage{float}

\def\BibTeX{{\rm B\kern-.05em{\sc i\kern-.025em b}\kern-.08em
    T\kern-.1667em\lower.7ex\hbox{E}\kern-.125emX}}

\begin{document}

\title{Unified Planning–Learning Framework for Robust UUV Navigation Under Partial Observability}


\author{
Md Ether Deowan$^{1}$ and Eleni Kelasidi$^{1}$%
\thanks{$^{1}$Department of Mechanical and Industrial Engineering, Faculty of Engineering, NTNU, Trondheim, Norway.
{\tt\small md.e.deowan@ntnu.no, eleni.kelasidi@ntnu.no}}
}

\maketitle
\thispagestyle{empty}
\pagestyle{empty}

\begin{abstract}

This paper presents an observation-only autonomy framework for Unmanned Underwater Vehicles (UUVs) navigation in dynamic underwater environments that integrates persistent occupancy mapping, global clearance-aware planning, and risk-aware local control. The proposed pipeline constructs occupancy maps solely from onboard sonar and depth image observations, adapts a clearance-constrained global planner (GP) to provide long-horizon structure, and integrates a reinforcement learning (RL) policy to handle short-range tracking and reactive avoidance. To further support decision-making under partial observability, the system learns a compact latent state representation from onboard sensor data, encoding environmental structure, obstacle dynamics, and uncertainty. Behavior tree (BT) distillation with staged supervision is introduced to improve safety and training stability, while an uncertainty-calibrated distillation mechanism reweights teacher guidance using online latent-model uncertainty, emphasizing uncertain regimes during learning, with time-to-collision (TTC) and clearance cues remaining explicit in planning and local policy features. To demonstrate the efficacy of the framework, a reproducible multi-seed evaluation protocol is established in high-fidelity GPU-accelerated simulation using NVIDIA Isaac Sim, and performance is benchmarked against BT-only and standard RL baselines. The results obtained demonstrate improved robustness and safety under dynamic conditions, thus providing a general pipeline with a unified hybrid planning–learning architecture and a reproducible methodology for robust UUV autonomy under partial observability.

\end{abstract}


\section{Introduction}
\label{sec:introduction}
Autonomous underwater navigation presents unique challenges that distinguish it from terrestrial and aerial robotics due to constrained sensing capabilities, complex hydrodynamic effects, GPS-denied localization, and highly dynamic environments involving moving obstacles such as marine life and other vehicles \cite{Evjemo2026,zhu2022uuvsurvey,fossen2011handbook}. These factors lead to a partially observable and uncertain decision-making problem, where conventional motion planning approaches often exhibit limited performance \cite{zhu2022uuvsurvey}. Simultaneous Localization and Mapping (SLAM) methods have therefore been investigated for autonomous UUV operation \cite{Singh2025,Cardaillac2023}. However, current state-of-the-art systems such as SVIn2 and TURTLMap \cite{svin2,turtlmap} remain challenged in low-texture, repetitive, and unstructured underwater environments. More recent work on robust localization and mapping, such as the framework proposed in \cite{MarcoJob2025}, has demonstrated promising real-time applicability in dynamic aquaculture scenarios. Similarly, motion planning approaches in \cite{Xanthidis2023,Amundsen2024,Amundsen2024b} have enabled collision-free navigation in the presence of ocean currents, control errors, and dynamic environments containing static and dynamic obstacles. Nevertheless, these approaches rely on the assumption of known environments and predefined obstacle information.



Classical underwater navigation pipelines are largely built on graph-search and sampling-based planners such as A*, RRT, and RRT*, as well as clearance-aware methods such as Voronoi-based path planning, typically combined with reactive local obstacle avoidance\cite{zhu2022uuvsurvey,hart1968astar,lavalle1998rapidly,karaman2011rrtstar, choset2000sensor}. These approaches offer geometric interpretability and completeness guarantees, and remain dominant in UUV systems due to their clear structure and feasibility properties. However, they remain brittle under dynamic changes, uncertain perception, and time-varying scenes \cite{zhu2022uuvsurvey,fossen2011handbook}. Reactive controllers also lack strategic foresight and may produce suboptimal or unsafe behaviors, while the separation between perception, planning, and control can lead to inconsistent decisions when environmental assumptions are violated \cite{zhu2022uuvsurvey}. In dynamic environments, precomputed routes can quickly become invalid, and frequent replanning under noisy maps may introduce instability or excessive computational load \cite{zhu2022uuvsurvey,fossen2011handbook}. Recent underwater works attempt to improve reactivity by combining artificial potential fields with deep RL controllers, including CAPF-TD3 and PPO-IIFDS variants for 3D dynamic avoidance \cite{li2025capftd3,liu2025ppoiifds}, as well as reward-shaped TD3 variants targeting trajectory smoothness and energy efficiency \cite{su2026improvedtd3,chen2024deepseaminingdrl}. Although these methods enhance local responsiveness, they are commonly evaluated under simplified obstacle models or constrained scenario assumptions, often without strict observation-only sensing constraints and mapping uncertainty.

A related limitation appears at the perception layer. Occupancy grid mapping (OGM) remains the standard representation for fusing uncertain range measurements into probabilistic free/occupied space \cite{elfes1989ogm}. In underwater robotics, OGMs are widely paired with forward-looking sonar to enable real-time obstacle detection and navigation. Recent work demonstrates OGM-based obstacle avoidance coupled with potential-field control \cite{jin2024ogm}, and sonar-based SLAM systems integrating filtering-based state estimation \cite{mu2022ogmslam}. While effective for local obstacle handling, many OGM-based systems remain predominantly reactive and short-horizon. Their robustness can degrade when dynamic obstacles appear abruptly or when sonar artifacts such as multipath and bubbles corrupt map updates. Moreover, mapping, planning, and control are frequently evaluated in isolation, masking interface-level failure modes that arise when perception uncertainty propagates into planning decisions.

To improve adaptability under such uncertainty, deep reinforcement learning (DRL), particularly model-free algorithms such as Proximal policy optimization (PPO) and Soft Actor-Critic (SAC) \cite{schulman2017ppo,haarnoja2018sac, 11127836}, offers a promising alternative by learning end-to-end policies that map observations directly to actions. These methods have been increasingly applied to underwater control tasks including station-keeping, tracking, and docking \cite{schulman2017ppo,haarnoja2018sac}. However, pure DRL approaches remain sample-intensive, exhibit unsafe exploration during early training, and often assume privileged state access or simplified obstacle models \cite{marinegym2025}. As a result, many learned policies are evaluated under conditions that do not reflect observation-only deployment constraints.

Because of this sample complexity, high-throughput simulation platforms play a central role in underwater robot learning. MarineGym has significantly improved training throughput and reproducibility through GPU-accelerated hydrodynamics and standardized benchmarks \cite{marinegym2025}. In parallel, perception-focused simulators provide increasingly realistic underwater sensing models, including imaging sonar and physics-based rendering, helping reduce the sim-to-real gap for observation-driven autonomy \cite{oceansim2025,potokar2022holoocean,amer2023unavsim,manhaes2016uuvsim,grimaldi2025stonefish}. Representative platforms include OceanSim \cite{oceansim2025}, HoloOcean \cite{potokar2022holoocean}, UNav-Sim \cite{amer2023unavsim}, UUV Simulator \cite{manhaes2016uuvsim}, and Stonefish \cite{grimaldi2025stonefish}. These advances enable more realistic evaluation, but they do not by themselves resolve the architectural gap between mapping, planning, and safety-aware learning.

This gap motivates hybrid planning--learning architectures that combine the long-horizon structure of classical planners with the adaptability of learned local controllers \cite{zhu2022uuvsurvey}. Policy distillation and imitation learning transfer structured priors from expert or rule-based controllers into neural policies \cite{rusu2016policydistillation}, while BTs provide interpretable and supervisory logic widely used in robotics autonomy stacks \cite{colledanchise2018bt}. However, existing distillation approaches typically apply fixed supervision schedules and uniform weighting of teacher signals, without conditioning guidance strength on real-time collision risk or environmental uncertainty. As a result, safety-critical regimes such as near-obstacle interactions or replanning transitions are not explicitly emphasized during learning.

Overall, prior work has advanced individual components of underwater autonomy, including global planning, sonar-based mapping, RL control, and high-fidelity simulation, but often treats them in isolation or under partially privileged settings. A persistent gap remains in unified, observation-only autonomy stacks that tightly couple persistent sonar-based mapping, clearance-aware global replanning, and risk-aware learned local control under moving obstacles.

This paper proposed a hybrid architecture that combines the complementary strengths of classical planning and learned decision-making, and thus addressed this integration gap by enforcing observation-only constraints end-to-end and introducing risk-calibrated supervision to stabilize safety-critical learning in dynamic environments. Therefore, unlike prior work that evaluates isolated modules, the developed framework enforces observation-only constraints across mapping, planning, and control, enabling realistic assessment of full autonomy stacks under dynamic uncertainty.

In summary, the contributions of this work are:
\begin{itemize}
    \item[C1.] A hierarchical motion planning framework integrating clearance-aware Voronoi-based global planning with RRT fallback and learned local policies for UUV navigation in dynamic underwater environments.
    \item[C2.] A BT distillation method that bootstraps deep RL training with expert demonstrations, significantly improving sample efficiency and safety.
    \item[C3.] A latent world model architecture that compactly represents environmental structure, obstacle dynamics, and uncertainty from partial sensor observations.
    \item[C4.] Extensive validation in high-fidelity GPU-accelerated simulation demonstrating successful navigation in complex underwater environments with curriculum learning.
    \item[C5.] Integration of sonar-based perception with occupancy grid mapping for real-time dynamic obstacle handling.
\end{itemize}

The remainder of this paper is organized as follows: Section~\ref{sec:methodology} details the proposed framework including the latent world model, hierarchical planning architecture, and BT distillation approach. Section~\ref{sec:experiments} describes the experimental setup and training procedures. Section~\ref{sec:results} presents quantitative results and analysis. Section~\ref{sec:conclusion} concludes with discussion and future work.

\section{Methodology}
\label{sec:methodology}
\begin{figure}[H]
    \centering
    \includegraphics[width=0.850\linewidth,trim=19pt 140pt 19pt 28pt,clip]{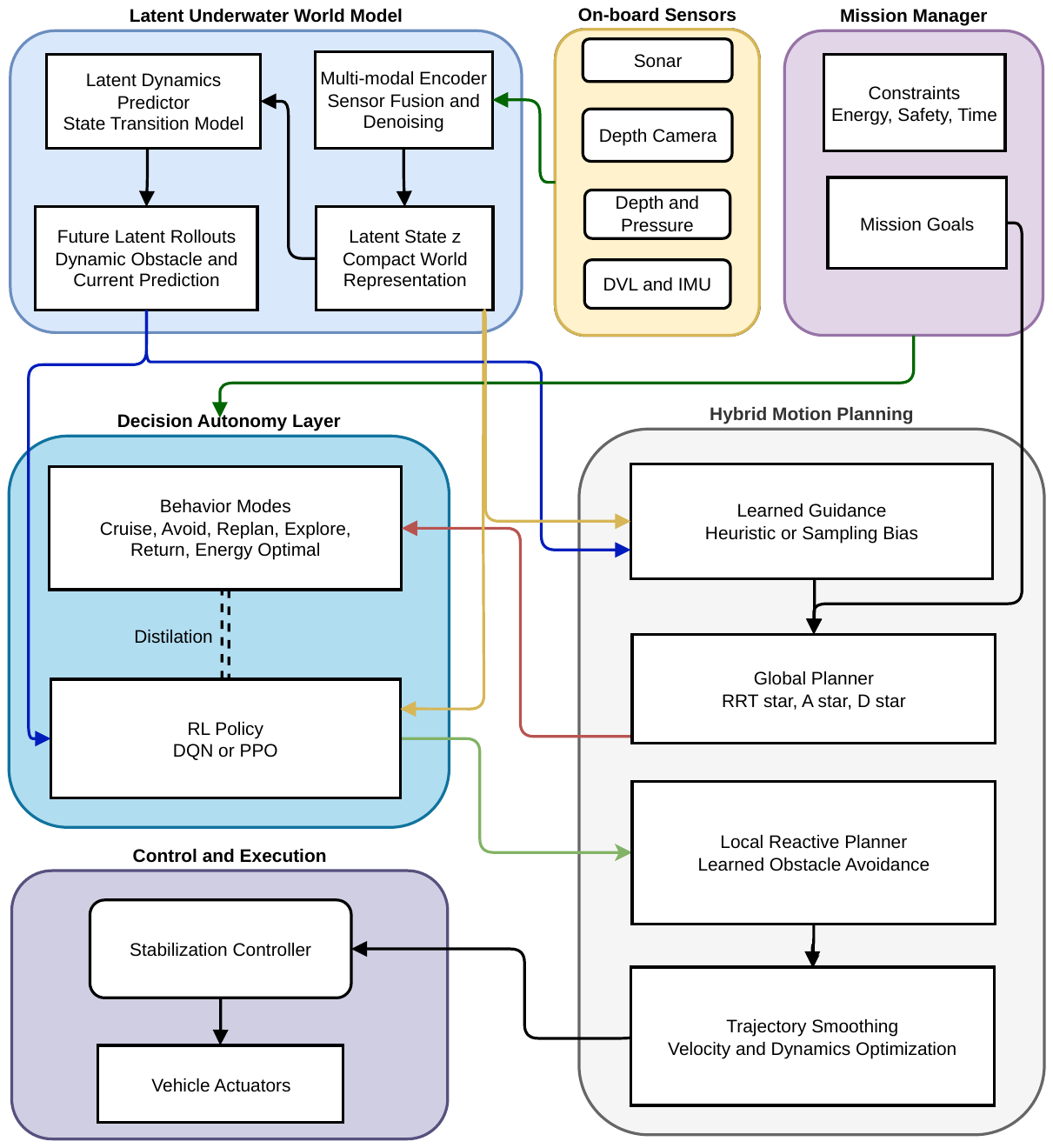}
    \caption{Latent world model guided hybrid autonomy architecture. Multi-modal sensing is fused into a compact latent state with learned dynamics rollouts for risk-aware prediction, which informs learned guidance, global and local planning, RL-based behavior selection, and closed-loop control.}
    \label{fig:architecture}
    \vspace{-20pt}
\end{figure}

\begin{figure*}[t]
    \centering
    \includegraphics[width=01.0\linewidth]{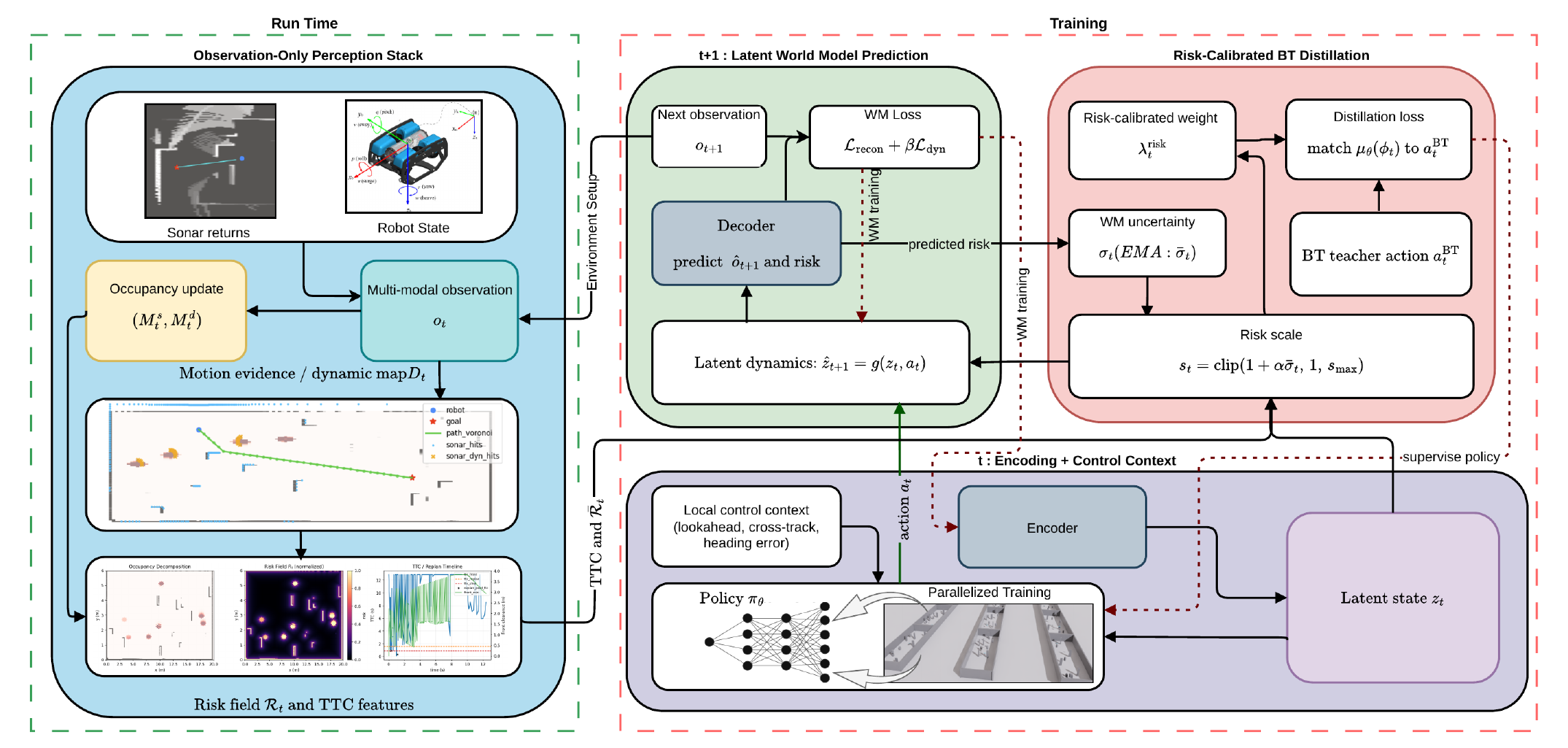}
    \caption{Overview of the proposed observation-only autonomy framework. Multi-modal sonar and state inputs are fused into a latent representation $z_t$. A learned latent dynamics model predicts $\hat{z}_{t+1}$ and $\hat{o}_{t+1}$, producing uncertainty $\hat{\sigma}_t$ and online residual features appended to policy observations. TTC and path-risk signals are used in local policy/planning features, while BT distillation is uncertainty-calibrated through $\lambda_t^{\text{risk}}$. The policy outputs continuous control actions $a_t$ and operates in a closed-loop perception--prediction--control pipeline.}
    \label{fig:autonomy_framework}
    \vspace{-15pt}
\end{figure*}

The navigation is formulated as a partially observable stochastic control problem with latent physical state $x_t$, onboard observation $o_t$, and action $a_t=[u_t,v_t,w_t,r_t]$ (surge, sway, heave, yaw-rate). The system evolves as
\begin{equation}
x_{t+1}\sim p(x_{t+1}\mid x_t,a_t,\omega_t),\qquad
o_t\sim p(o_t\mid x_t,\nu_t),
\end{equation}
with process disturbance $\omega_t$ and sensor noise $\nu_t$. The policy is optimized under strict observation-only constraints:
\begin{equation}
o_t=\left[o_t^{\mathrm{sonar}},o_t^{\mathrm{depth}},o_t^{\mathrm{imu}},o_t^{\mathrm{vel}},o_t^{\mathrm{path}}\right]
\end{equation}
and objective
\begin{equation}
\begin{aligned}
\max_{\pi_\theta}\;&\mathbb{E}_{\pi_\theta}\!\left[\sum_{t=0}^{T}\gamma^t r_t\right]\\[-1mm]
\text{s.t.}\;& \Pr(\text{collision})\le \epsilon,\; a_t\in\mathcal{A},\; x_t\in\mathcal{X},\; \pi_\theta(a_t|o_{\le t}).
\end{aligned}
\end{equation}
Here, $\mathbb{E}_{\pi_\theta}$ denotes expectation over trajectories induced by policy $\pi_\theta$, $r_t$ is reward, $T$ is the horizon, $\gamma$ is the discount factor, $\epsilon$ is the collision-risk bound, and $\mathcal{A}$/$\mathcal{X}$ encode actuator/feasibility limits for $a_t$/$x_t$.
The end-to-end autonomy stack is a coupled recursion
\begin{equation}
\begin{aligned}
\left(M_t^{\mathrm{s}},M_t^{\mathrm{d}},\Pi_t,\phi_t,a_t\right)
=
\mathcal{G}\!\Big(&M_{t-1}^{\mathrm{s}},M_{t-1}^{\mathrm{d}},\\
&\Pi_{t-1},o_t,a_{t-1}\Big),
\end{aligned}
\end{equation}
where $M_t^{\mathrm{s}}$ and $M_t^{\mathrm{d}}$ are static and dynamic occupancy memories, $\Pi_t$ is the global path, and $\phi_t$ are local-policy features. The operator $\mathcal{G}(\cdot)$ denotes the closed-loop update map of the full autonomy stack. As illustrated in Fig.~\ref{fig:architecture}, persistent mapping and risk accumulation correspond to Sec.~\ref{subsec:mapping_risk}; latent world modeling and uncertainty estimation correspond to Sec.~\ref{subsec:latent_uncertainty}; and global planning, local decision-making, and closed-loop control correspond to Sec.~\ref{subsec:planning_control}. The BT distillation module in Sec.~\ref{subsec:bt_distill} is used during training to shape the policy and which is shown in Fig.~\ref{fig:autonomy_framework}.

\subsection{Sonar-Based Mapping and Risk Representation}
\label{subsec:mapping_risk}
Occupancy is updated from sonar hit/free-space evidence and depth-gated consistency. The implementation maintains separate static and dynamic layers:
\begin{equation}
\begin{aligned}
M_t^{\mathrm{s}}(c)=\operatorname{clip}\!\Big(
&(1-\lambda_s)M_{t-1}^{\mathrm{s}}(c)
+\eta_{\mathrm{occ}}I_t^{\mathrm{s\text{-}hit}}(c)\\
&-\eta_{\mathrm{free}}I_t^{\mathrm{free}}(c)
+\eta_{\mathrm{depth}}I_t^{\mathrm{depth}}(c),\,0,1
\Big),
\end{aligned}
\end{equation}
\begin{equation}
\begin{aligned}
M_t^{\mathrm{d}}(c)=&\operatorname{clip}\!\Big(
(1-\lambda_d)M_{t-1}^{\mathrm{d}}(c)
+\eta_{\mathrm{dyn}}I_t^{\mathrm{d\text{-}hit}}(c)\\
&-\eta_{\mathrm{dyn,free}}I_t^{\mathrm{free}}(c)
+\eta_{\mathrm{stamp}}I_t^{\mathrm{stamp}}(c),\,0,1
\Big),
\end{aligned}
\end{equation}
with fused occupancy
\begin{equation}
M_t(c)=\max\!\left(M_t^{\mathrm{s}}(c),M_t^{\mathrm{d}}(c)\right).
\end{equation}
Each layer stores a bounded cell-wise occupancy belief in $[0,1]$ where $c$ indexes a grid cell in the local occupancy map. Here, $\lambda_s,\lambda_d$ are static/dynamic decay factors, and $\eta_{\mathrm{occ}},\eta_{\mathrm{free}},\eta_{\mathrm{depth}},\eta_{\mathrm{dyn}},\eta_{\mathrm{dyn,free}},\eta_{\mathrm{stamp}}$ are bounded update gains. The terms $I_t^{\mathrm{s\text{-}hit}}(c)$, $I_t^{\mathrm{d\text{-}hit}}(c)$, $I_t^{\mathrm{free}}(c)$, $I_t^{\mathrm{depth}}(c)$, and $I_t^{\mathrm{stamp}}(c)$ are binary indicators for static-hit, dynamic-hit, free-space, depth-consistency, and dynamic-stamp evidence, respectively. During runtime, dynamic occupancy is inferred from observation-derived motion evidence only (temporal sonar inconsistency and Doppler-relative motion). Simulator-side dynamic stamping is used only as an optional debugging aid, meaning that all reported experiments keep this term disabled to preserve strict observation-only sensing. Planning uses an online clearance field
\begin{equation}
\delta_t(c)=\min\!\left(d_{\mathrm{occ},t}(c),d_{\mathrm{bnd}}(c)\right),
\end{equation}
where $d_{\mathrm{occ},t}(c)$ is distance-to-occupied space at time $t$ and $d_{\mathrm{bnd}}(c)$ is distance-to-boundary. For path diagnostics, a scalar risk map and its path-averaged density are used:
\begin{equation}
\mathcal{R}_t(c)=\alpha_m M_t^{\mathrm{s}}(c)+\alpha_d M_t^{\mathrm{d}}(c)+\alpha_\delta\frac{1}{\delta_t(c)+\varepsilon}
\end{equation}
\begin{equation}
\bar{\mathcal{R}}_t=\frac{1}{H}\sum_{j=1}^{H}\mathcal{R}_t(p_{t,j}),
\end{equation}
where $\alpha_m,\alpha_d,\alpha_\delta$ are weight static occupancy, dynamic occupancy, and clearance penalty, respectively; $p_{t,j}$ is the $j$th point on the finite-horizon lookahead path segment $\Pi_t^{(H)}=\{p_{t,1},\ldots,p_{t,H}\}$. Here, $H$ is the lookahead horizon and $\varepsilon>0$ is a numerical regularizer.

\subsection{Hierarchical Global and Local Planning and Control}
\label{subsec:planning_control}

The global planner operates over the updated occupancy and clearance memories. Global planning is constrained to traversable cells
\begin{equation}
\Omega_t=\{c\mid M_t(c)<\tau_{\mathrm{occ}},\;\delta_t(c)\ge d_{\min}\},
\end{equation}
and path synthesis is posed as
\begin{equation}
P_t^\star=\arg\min_{P\subset\Omega_t}\sum_i
\frac{\|p_{i+1}-p_i\|_2}{\max(\delta_t(p_i),\varepsilon_r)}.
\end{equation}
Here, $P=\{p_i\}$ is a candidate waypoint sequence and $P_t^\star$ is the selected minimum-cost path; $\tau_{\mathrm{occ}}$ is the occupancy threshold, $d_{\min}$ is the minimum clearance, and $\varepsilon_r$ regularizes low-clearance costs. Voronoi planning is executed first because medial-axis paths maximize clearance in narrow passages and are less sensitive to local sonar-map uncertainty than shortest grid paths. A pure occupancy-grid A* planner was not used as the primary global planner because, without aggressive inflation, grid-optimal paths can hug occupied cells and amplify sonar/map discretization errors; RRT remains as a fallback when Voronoi connectivity fails. Replanning is event-triggered by: (i) no-progress/stuck detection, implemented as low speed away from the goal for a fixed window; (ii) safety/TTC events, using near-obstacle range and TTC termination thresholds; and (iii) map or dynamic-obstacle intersection with the lookahead path, including a constant-velocity forecast over a 2.5~s horizon.

The local policy $\pi_\theta$ receives path-relative and proprioceptive features
\begin{equation}
\phi_t=
\left[
\Delta p_t^{\mathrm{lh}},\,e_t^{\psi},\,e_t^{\mathrm{ct}},\,
v_t^{b},\,\omega_t^{b},\,
\xi_t^{\mathrm{sonar}},\,\xi_t^{\mathrm{wm}}
\right],
\end{equation}
where $\Delta p_t^{\mathrm{lh}}$ is lookahead displacement, $e_t^{\psi}$ is heading error, $e_t^{\mathrm{ct}}$ is cross-track error, $\xi_t^{\mathrm{sonar}}$ are sectorized range/TTC features, and $\xi_t^{\mathrm{wm}}$ are latent world-model features. Here, $v_t^{b}$ and $\omega_t^{b}$ denote body-frame linear and angular velocities, respectively. In practice, $\xi_t^{\mathrm{wm}}$ contains the one-step prediction residual and an uncertainty proxy derived from the latent world model. Reward shaping is defined as
\begin{align}
r_t &= r_{\mathrm{alive}} + w_p\Delta s_t - w_d d_t
- w_{\mathrm{ct}}|e_t^{\mathrm{ct}}|
- w_c C_t - w_{\mathrm{ttc}}T_t \nonumber\\
&\quad - w_v\|v_t\|_{\mathrm{over}}^2
+ w_u U_t + w_h H_t
+ w_s\iota_t^{\mathrm{succ}}
- w_f\iota_t^{\mathrm{fail}} \nonumber\\
&\quad - w_{\mathrm{tilt}}\Theta_t - w_{\mathrm{eff}}E_t,
\end{align}
where $\Delta s_t$ is path progress, $d_t$ is path-reference distance, $C_t$ is soft clearance violation, $T_t$ is TTC penalty, and $\|v_t\|_{\mathrm{over}}$ penalizes overspeed beyond configured safety velocity. Here, $r_{\mathrm{alive}}$ is a constant per-step survival reward, $U_t$ denotes uprightness, and $H_t$ denotes heading-alignment reward; $\Theta_t$ is tilt magnitude and $E_t$ is mean control effort. Terminal indicators are defined as $\iota_t^{\mathrm{succ}}\in\{0,1\}$ and $\iota_t^{\mathrm{fail}}\in\{0,1\}$, equal to $1$ only when success or failure termination is triggered at step $t$, respectively. The coefficients $w_p,w_d,w_{\mathrm{ct}},w_c,w_{\mathrm{ttc}},w_v,w_u,w_h,w_s,w_f,w_{\mathrm{tilt}},w_{\mathrm{eff}}$ are manually selected from stable validation rollouts, then fixed for all PPO comparisons; no automated reward-weight search or sensitivity sweep is claimed. The reward combines progress tracking, safety penalties, attitude stabilization, and terminal success/failure terms. In addition to reward shaping, hard safety constraints are enforced online:
\begin{equation}
d_t^{\min}\ge d_{\mathrm{coll}},\qquad \mathrm{TTC}_t\ge \tau_{\mathrm{coll}},
\end{equation}
with immediate termination when violated, ensuring policy updates are always conditioned on explicit safety boundaries. Here, $d_t^{\min}$ is the minimum instantaneous obstacle clearance, $d_{\mathrm{coll}}$ is the collision-clearance threshold, and $\tau_{\mathrm{coll}}$ is the TTC safety threshold.

\subsection{BT-Guided Distillation for Policy Learning}
\label{subsec:bt_distill}

To improve sample efficiency and suppress unsafe exploration, RL is trained with BT distillation:
\begin{equation}
\mathcal{L}(\theta)=
\mathcal{L}_{\mathrm{RL}}(\theta)
+\lambda_t\,
\mathbb{E}\!\left[
\|\mu_\theta(\phi_t)-a_t^{\mathrm{BT}}\|_2^2
\right],
\end{equation}
where $\mathcal{L}_{\mathrm{RL}}$ denotes the underlying RL objective (e.g., PPO loss), and $a_t^{\mathrm{BT}}$ is the teacher action generated by the behavior tree. The staged schedule is
\begin{equation}
\lambda_t=
\begin{cases}
1.0, & t\le 800,\\[4pt]
0.2\left(1-\dfrac{t-800}{1600-800}\right), & 800<t\le 1600,\\[6pt]
0, & t>1600,
\end{cases}
\end{equation}
where $t$ denotes policy-update iteration. This schedule transitions from expert-guided stabilization for optimization.

\begin{figure}[t]
    \centering
    \includegraphics[width=0.90\linewidth,trim=38pt 200pt 28pt 34pt,clip]{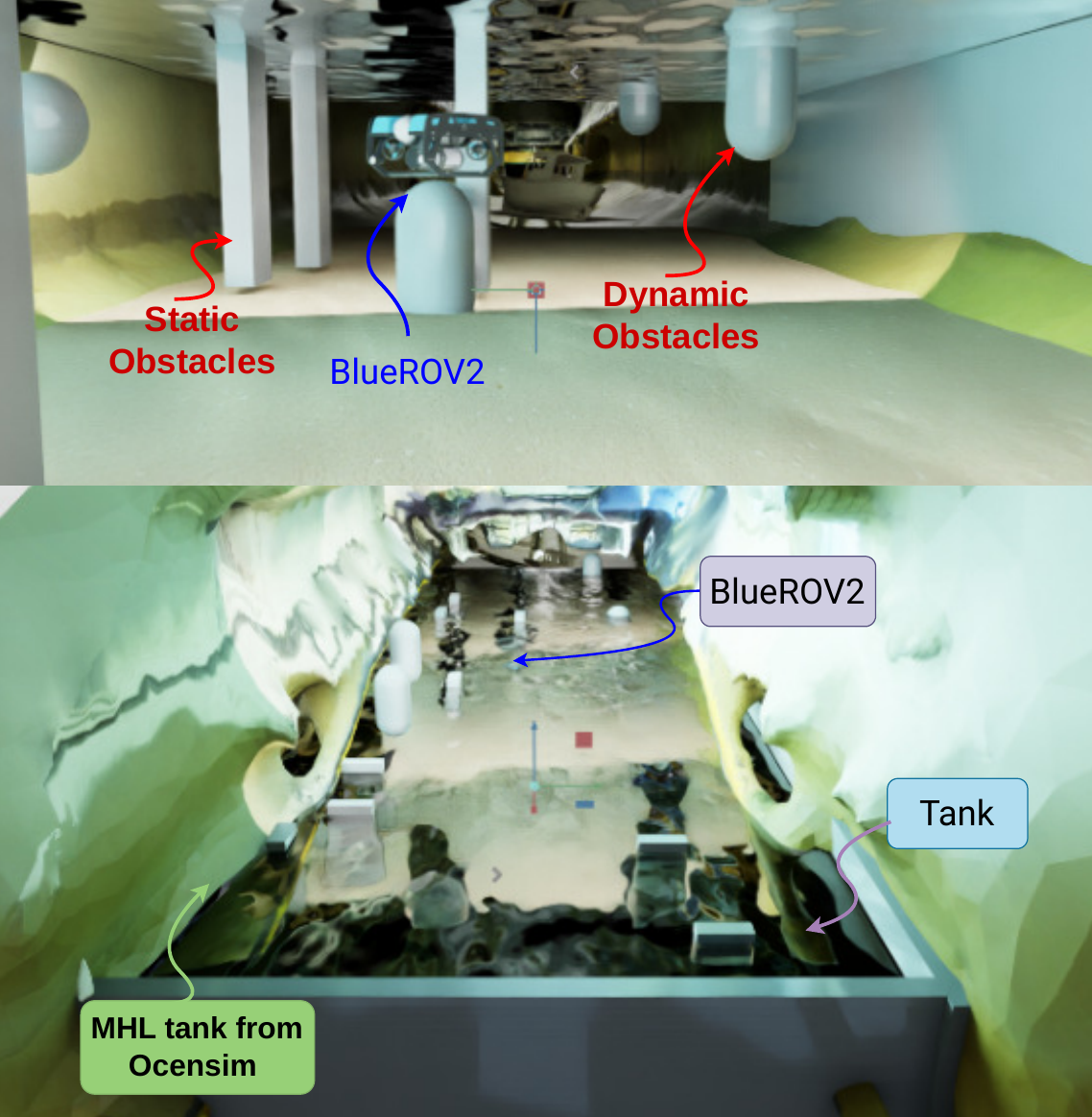}
    \caption{Representative simulation environment with dynamic obstacles. The scene contains a mix of static and moving obstacles, with the UUV tasked to navigate from a start location to a goal while avoiding collisions. }
    \label{fig:sim_env}
    \vspace{-10pt}
\end{figure}

\subsection{Latent World Model and Uncertainty-Aware Supervision}
\label{subsec:latent_uncertainty}

Partial observability is addressed with an online latent predictor:
\begin{align}
z_t&=f_\varphi(o_t),\enspace
\hat{z}_{t+1}=g_\psi(z_t,a_t),\enspace
\hat{o}_{t+1}=h_\omega(\hat{z}_{t+1}),
\end{align}
where $o_t$ and $a_t$ are the observation and action, and $f_\varphi$, $g_\psi$, and $h_\omega$ denote the encoder, latent dynamics model, and decoder. The loss has two parts: one asks the decoder to predict selected next-observation terms, and the other asks the dynamics model to predict the next latent state:
\begin{equation}
\mathcal{L}_{\mathrm{WM}}
=
\mathcal{L}_{\mathrm{recon}}(\hat{o}_{t+1},\tilde{o}_{t+1})
\;+\;
\beta \|z_{t+1}-\hat{z}_{t+1}\|_2^2,
\end{equation}
where $\beta$ weights the latent-dynamics loss. Both targets are one-step quantities: $\tilde{o}_{t+1}=o_{t+1}^{(1:d_{\mathrm{tar}})}$ is the first $d_{\mathrm{tar}}$ entries of the next observation, and $z_{t+1}=f_\varphi(o_{t+1})$ is the encoded next observation. Here $d_{\mathrm{tar}}=24$: 3 relative goal-position errors in the body frame, 2 heading-error terms $[\sin e_t^{\psi},\cos e_t^\psi]$, 3 body-frame velocities, yaw rate, depth, roll and pitch, and the first 12 sonar ranges. World-model (WM) architecture, optimizer settings, and $\beta$ are listed in Table~\ref{tab:app_repro}. $\mathcal{L}_{\mathrm{recon}}$ is Gaussian negative log likelihood (NLL), or mean-squared error (MSE) when NLL is disabled; the decoder log-variance $\log\widehat{\mathrm{var}}_t$ provides an uncertainty proxy
\begin{equation}
\sigma_t=\operatorname{mean}\!\big(\exp(\log\widehat{\mathrm{var}}_t)\big),
\end{equation}
which is tracked by an exponential moving average
\begin{equation}
\bar{\sigma}_t=(1-\rho)\bar{\sigma}_{t-1}+\rho \sigma_t,
\end{equation}
where $\rho\in(0,1]$ is the EMA update factor. This quantity is a decoder-variance proxy used for policy-feature augmentation and distillation scaling; it is not a separately calibrated collision probability or probabilistic safety guarantee.

The policy receives two online WM features (prediction residual and uncertainty proxy), and BT distillation is uncertainty-calibrated by
\begin{equation}
s_t=\operatorname{clip}\!\left(1+\alpha\bar{\sigma}_t,\;1,\;s_{\max}\right),\qquad
\lambda_t^{\mathrm{risk}}=\lambda_t s_t,
\end{equation}
where $\lambda_t$ is the staged base distillation weight, $\alpha\!\ge\!0$ controls uncertainty-to-supervision gain, and $s_{\max}\!\ge\!1$ bounds multiplicative amplification. The gains are given in Table~\ref{tab:app_repro}; they keep risk scaling bounded while increasing teacher influence in high-uncertainty rollouts.
The mechanism's runtime/training separation and data flow are illustrated in Fig.~\ref{fig:autonomy_framework}.

\subsection{Ablation Protocol and Implementation}
\label{subsec:ablation_protocol}

To isolate the effect of each architectural component, a controlled ablation set is evaluated under identical seeds and scenario distributions: (A) \textit{Full}: OGM + GP + BT curriculum distillation + LWM with uncertainty scaling ($\alpha>0$); (B) \emph{LWM risk}: same as Full with $\alpha=0$; (C) \emph{LWM}: world model input removed ($\xi_t^{\mathrm{wm}}=\mathbf{0}$); (D) \emph{BT distill}: pure PPO under the same mapping and planning stack. To validate the hierarchical planner, all variants are evaluated using success rate, collision rate, episodic return, average episode time, and time-to-success (TTS). The corresponding aggregate performance comparison is reported in Table~\ref{tab:main}.

\begin{table}[H]
\centering
\caption{Navigation performance in underwater scenarios. Lower TTS indicates faster and more reliable navigation.}
\label{tab:main}
\scriptsize
\setlength{\tabcolsep}{2.0pt}
\renewcommand{\arraystretch}{0.88}
\begin{tabular*}{\columnwidth}{@{\extracolsep{\fill}}lccccc@{}}
\toprule
Method & Succ. $\uparrow$ & Coll. $\downarrow$ & Return $\uparrow$ & Avg. (s) $\downarrow$ & TTS (s) $\downarrow$ \\
\midrule
PPO w/o distill        & 0.20 & 0.40  & +197.3  & 22.84 & 114.2 \\
PPO 20\% distill       & 0.46 & 0.20  &  +173.4 & 13.44 & 67.2  \\
PPO curr.\ distill     & 0.87 & 0.15  & +176.4  & 11.41 & 15.8  \\
BT                     & 0.80 & 0.10  & +40.0    & 14.50 & 32.5  \\
\bottomrule
\end{tabular*}
\vspace{-15pt}
\end{table}

Each control cycle performs: (i) map and risk updates $(M_t^{\mathrm{s}}, M_t^{\mathrm{d}})$, (ii) global planning $\Pi_t$ via Voronoi with event-triggered RRT fallback, (iii) local control $a_t=\pi_\theta(\phi_t)$, and (iv) safety termination checks. Training is conducted in vectorized Isaac Sim with safety penalties active. Reward design, termination conditions, and evaluation protocol remain fixed across ablations where only the targeted component is modified, ensuring attributable performance differences.

\section{Experimental Setup}
\label{sec:experiments}

All experiments are conducted in Isaac Sim with GPU-accelerated underwater physics in a pool-like maze with static and moving obstacles. The simulated platform is a BlueROV2-class vehicle (MarineGym BlueROV asset) \cite{marinegym2025}; key hydrodynamic, sensing, timing, and safety parameters are listed in Table~\ref{tab:app_repro}. The vehicle executes continuous 4-DoF commands with pitch and roll stabilization $a_t=[u_t,v_t,w_t,r_t]$. Policy inputs are strictly observation-only, using sonar-derived, proprioceptive, and planner-relative path features as discussed in Sec.~\ref{sec:methodology}. Measured workstation timing is summarized in Table \ref{tab:latency}.

For learning-based methods, training is implemented through the Isaac Lab reinforcement learning stack \cite{isaaclab2024}. The primary learner is PPO, trained using the RL-Games and RSL-RL backends; PPO and auxiliary-model settings are summarized in Table~\ref{tab:app_repro}. BT supervision follows the staged distillation schedule in Sec.~\ref{sec:methodology}. In world-model runs, two online features, uncertainty and one-step prediction residual, are appended to the policy input. Safety is enforced during both training and evaluation via collision/TTC termination. Dynamic obstacles follow scripted kinematic waypoint/loop trajectories with randomized phase and assigned nominal speed; they do not actively avoid or react to the UUV. The robustness sweep summarized in Fig.~\ref{fig:policy_sweep} assigns the same nominal speed $v$ to all moving obstacles in a condition, abstracting more complex marine-life and current-driven motion. Robustness is encouraged through randomized start/goal placement, obstacle layouts, and interaction conditions; hydrodynamic parameters are kept fixed to the active vehicle configuration. The OceanSim check in Fig.~\ref{fig:sim_env} uses a separate pool-style scene, rendering pipeline, and scene geometry, and should be interpreted as a qualitative perception/scene-transfer check rather than a full hydrodynamic transfer benchmark.

\begin{table}[t]
\centering
\caption{Workstation deployability timing. Values are not embedded-hardware guarantees.}
\label{tab:latency}
\scriptsize
\setlength{\tabcolsep}{2.0pt}
\renewcommand{\arraystretch}{0.88}
\begin{tabular*}{\columnwidth}{@{\extracolsep{\fill}}lcc@{}}
\toprule
Component & Mean & Max / note \\
\midrule
WM encoder & 0.015 ms & CPU microbench. \\
WM dynamics & 0.015 ms & CPU microbench. \\
WM decoder & 0.027 ms & mean+logvar \\
PPO policy & 0.110 ms & 0.127 ms \\
Mapping update & 3.33 ms & 6.17 ms \\
Global planner & 5.22 ms & 9.89 ms \\
Full Python/Isaac loop & 28.63 ms & 39.88 ms; 20 ms budget \\
Replanning rate & 6.21 Hz & 52.1 replans/episode \\
\bottomrule
\end{tabular*}
\vspace{-15pt}
\end{table}

\begin{table}[H]
\centering
\caption{Runtime versus dynamic-obstacle density.}
\label{tab:runtime_density}
\scriptsize
\setlength{\tabcolsep}{2.0pt}
\renewcommand{\arraystretch}{0.88}
\begin{tabular*}{\columnwidth}{@{\extracolsep{\fill}}lccccc@{}}
\toprule
$D$ obstacles & 0 & 3 & 6 & 9 & 15 \\
\midrule
Loop time (ms) & 19.7 & 25.2 & 28.6 & 31.6 & 38.0 \\
Replans/episode & 51.9 & 62.7 & 54.4 & 48.8 & 42.7 \\
\bottomrule
\end{tabular*}
\vspace{-10pt}
\end{table}
Experiments are run on an AMD Ryzen 9 9950X3D CPU, NVIDIA RTX 5090 (32~GB VRAM), and 60~GB RAM. Table~\ref{tab:latency} shows that learned inference is negligible relative to mapping/planning and simulator overhead; dense scenes can exceed the 50~Hz budget, so embedded deployment requires platform-specific profiling.

\section{Results and Analysis}
\label{sec:results}

\subsection{Training Efficiency and World Model Learning}

\begin{figure}[t]
    \centering
    \begin{subfigure}[t]{0.49\linewidth}
        \centering
        \includegraphics[width=\linewidth,height=0.12\textheight,keepaspectratio,trim=190pt 300pt 195pt 320pt,clip]{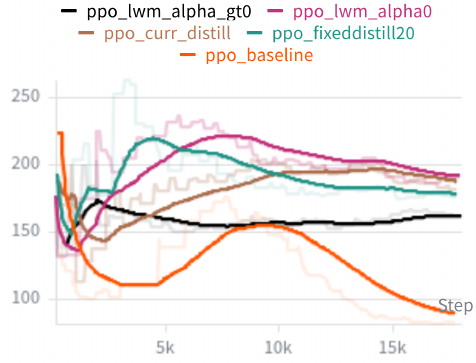}
        \caption{Mean training reward}
        \label{fig:mean_rewards}
    \end{subfigure}
    \hfill
    \begin{subfigure}[t]{0.49\linewidth}
        \centering
        \includegraphics[width=\linewidth,height=0.12\textheight,keepaspectratio,trim=175pt 285pt 175pt 310pt,clip]{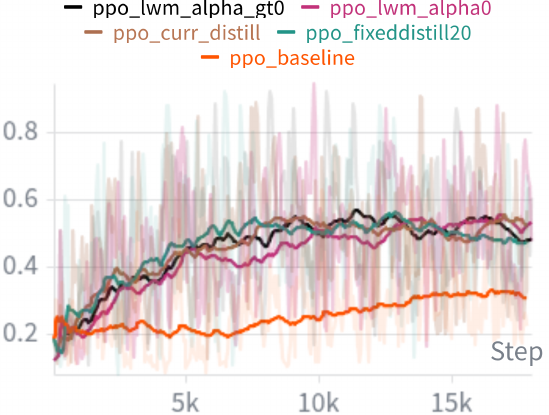}
        \caption{Task success rate}
        \label{fig:success_rate}
    \end{subfigure}

    \vspace{0.02cm}

    \begin{subfigure}[t]{0.49\linewidth}
        \centering
        \includegraphics[width=\linewidth,height=0.12\textheight,keepaspectratio,trim=175pt 285pt 175pt 310pt,clip]{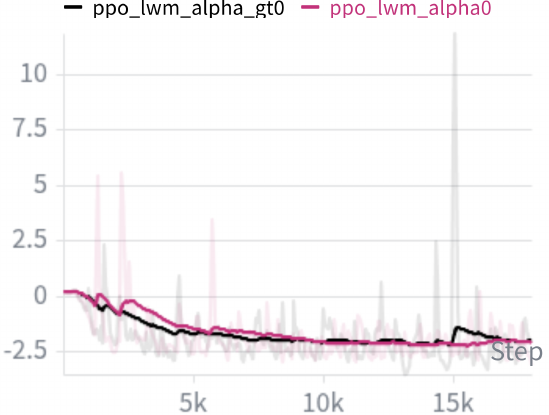}
        \caption{Reconstruction loss}
        \label{fig:recon_loss}
    \end{subfigure}
    \hfill
    \begin{subfigure}[t]{0.49\linewidth}
        \centering
        \includegraphics[width=\linewidth,height=0.12\textheight,keepaspectratio,trim=175pt 285pt 175pt 310pt,clip]{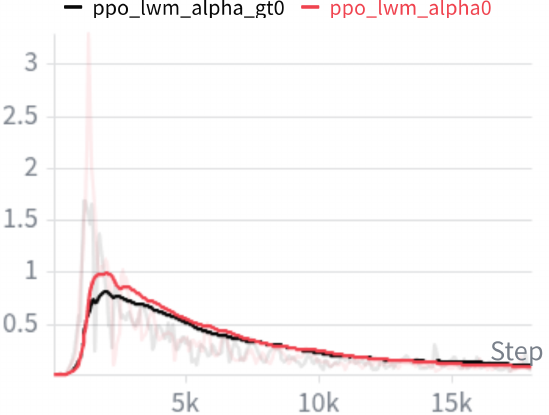}
        \caption{Dynamics prediction loss}
        \label{fig:dyn_loss}
    \end{subfigure}

    \caption{
    Training dynamics and latent world model convergence.
    Distillation-based variants exhibit accelerated learning, improved stability, and consistent one-step world model convergence compared to PPO without distillation.
    Training success is measured online during learning and is separate from the fixed multi-seed evaluation protocol in Table~\ref{tab:main}.
    }
    \label{fig:training_results}
    \vspace{-10pt}
\end{figure}

Fig.~\ref{fig:training_results} demonstrates that incorporating latent world modeling with distillation substantially improves learning dynamics. 
Distillation-based policies converge faster and achieve higher stable reward compared to PPO without distillation. 
The online training success curve is conservative because it is measured during exploratory rollouts, whereas Table~\ref{tab:main} reports fixed-policy, multi-seed evaluation without exploration noise. Crucially, reward improvements correspond to consistent increases in task success, confirming that gains are not attributable to reward shaping artifacts.

Simultaneously, reconstruction and dynamics losses decrease smoothly and stabilize, verifying that the latent world model learns compact representations and stable one-step predictive dynamics. 
This stable predictive structure provides a principled mechanism for improved policy updates and underpins the observed performance gains.

\begin{figure*}[t]
    \centering
    \includegraphics[width=\textwidth]{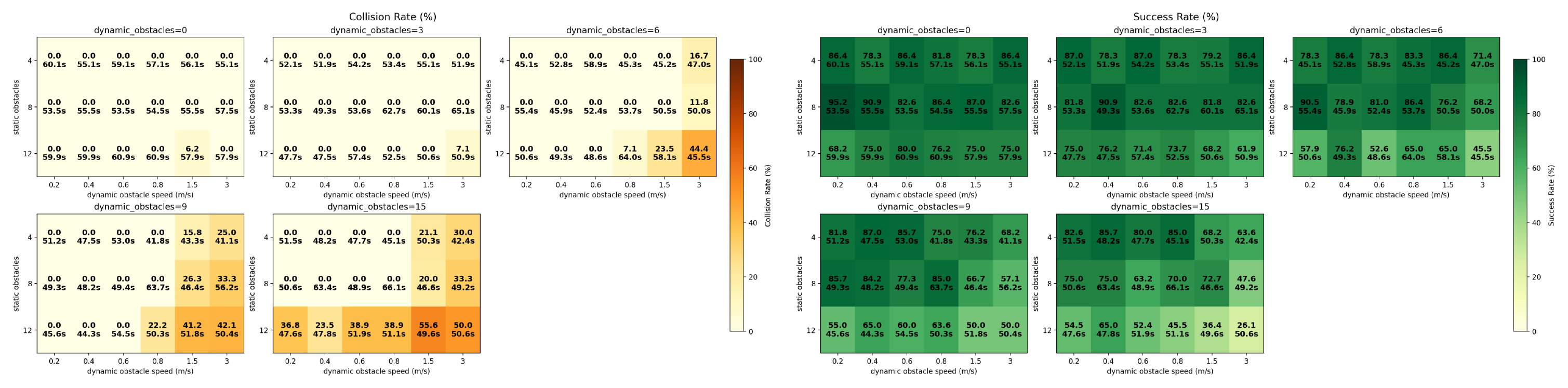}
    \caption{Robustness evaluation across static obstacle ($S=4,8,12$), dynamic obstacle counts ($D=0,3,6,9,15$), and speeds ($v=0.2$--$3.0$~m/s). Left: collision rate (\%). Right: success rate (\%). In-cell values report mean time-to-success (seconds).}
    \label{fig:policy_sweep}
    \vspace{-10pt}
\end{figure*}

\subsection{Quantitative Navigation Performance}

All methods are evaluated using the multi-seed protocol described in Sec.~\ref{sec:experiments} (5 seeds $\times$ 5 trials, 25 episodes per method). 
The reported metrics are success rate, collision rate (including TTC-based terminations), episodic return, average episode time, and time-to-success (TTS):

\begin{equation}
\mathrm{TTS}=\frac{\mathrm{AvgTime}}{\mathrm{SuccessRate}},
\end{equation}

which jointly penalizes slow execution and unreliability.

Table~\ref{tab:main} reveals a decisive performance advantage for the curriculum-distilled policy. 
It achieves nearly a 4$\times$ improvement in success rate over PPO without distillation, while simultaneously reducing collision rate and dramatically lowering TTS (15.8s vs. 114.2s). 
Episodic return is therefore treated as a diagnostic training signal rather than the primary operational metric. The higher return of PPO without distillation occurs because survival, progress, and attitude-alignment rewards can accumulate before a late collision or timeout, even when the episode does not complete the mission safely. For deployment-oriented comparison, success, collision, and TTS are prioritized because they directly measure goal completion, safety, and reliability-weighted speed.

The curriculum-distilled variant achieves a favorable safety--efficiency balance, approaching the robustness of the BT baseline while maintaining superior efficiency and adaptability. 
This demonstrates that latent world model distillation effectively regularizes policy learning and improves real-world operational reliability.
\begin{figure*}[t]
    \centering
    \begin{subfigure}{0.48\textwidth}
        \centering
        \includegraphics[width=\linewidth]{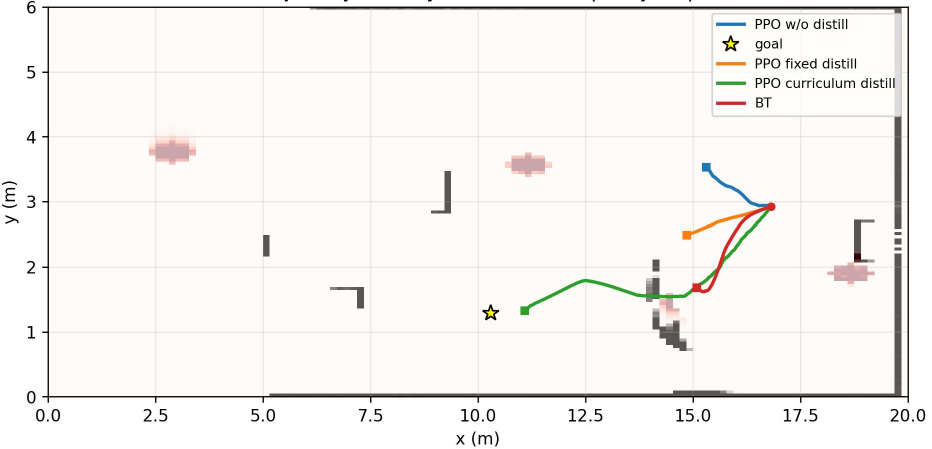}
        \caption{Scenario A}
        \label{fig:traj_case1}
    \end{subfigure}\hfill
    \begin{subfigure}{0.48\textwidth}
        \centering
        \includegraphics[width=\linewidth]{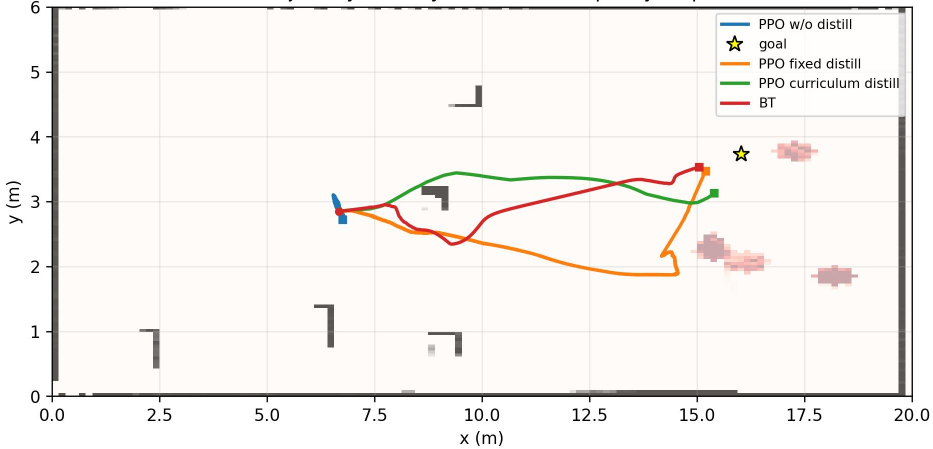}
        \caption{Scenario B}
        \label{fig:traj_case2}
    \end{subfigure}
    
    \caption{Representative trajectory rollouts overlaid on the sonar occupancy map. Gray/black cells denote occupied static-map evidence, red/pink overlays denote dynamic occupancy evidence, colored curves denote method trajectories, and the star marks the goal. Apparent visual overlap with occupied cells can occur from accumulated map projection and is not counted as collision unless the collision/TTC termination condition is triggered. Distillation-based policies produce smoother and more goal-directed trajectories with reduced oscillatory behavior.}
    \label{fig:trajectory_comparison}
    \vspace{-15pt}

\end{figure*}
\subsection{Robustness Under Increasing Dynamic Complexity}
\label{subsec:robustness_dynamic}

Fig.~\ref{fig:policy_sweep} provides a structured robustness test over jointly increasing scene complexity. Each condition is indexed by $(S,D,v)$, where $S\in\{4,8,12\}$ denotes static obstacle count with randomized layouts, $D\in\{0,3,6,9,15\}$ denotes dynamic obstacle count, and $v\in\{0.2,0.4,0.6,0.8,1.5,3.0\}$~m/s denotes dynamic obstacle speed. The left panel reports collision rate and the right panel reports success rate; each cell reports rate (\%) and mean time-to-success. 

Across low-to-moderate dynamic complexity, the policy sustains high success with low collision. For $S\le8$, $D\le9$, and $v\le1.5$~m/s, mean success is 82.8\% and mean collision is 1.1\%, with 38 of 40 conditions having zero collision. For $D\le3$, collision remains below 0.4\% on average over all static clutter and speed settings. This indicates that the observation-only stack remains stable under substantial variation in obstacle layout and motion.

Degradation appears mainly in the extreme corner cases: high static clutter, highest dynamic congestion, and fastest obstacles ($v=3.0$~m/s). This defines an operational boundary rather than a general failure mode: obstacle speed reaches 3.0~m/s while robot maximum speed is 0.6~m/s, reducing reaction margin and making evasive motion underactuated. The results also reflect a safety completion tradeoff: larger clearance/TTC margins can recover zero collision, but increase conservative detours, stuck episodes, and timeouts. Overall, the sweep shows robustness over practical conditions and graceful degradation under adversarial motion regimes.

\subsection{Qualitative Trajectory Analysis}

Fig.~\ref{fig:trajectory_comparison} complements Table~\ref{tab:main} by showing how the metrics appear in trajectory space. Scenario A represents a moderately cluttered case with local obstacle interaction near the goal region; Scenario B represents a harder case requiring larger heading correction and longer obstacle-aware detouring. In both scenarios, the non-distilled PPO policy exhibits less stable route selection and larger path inefficiency, while distillation-based policies are more directed and consistent.

The qualitative trends are aligned with the quantitative outcomes: curriculum-distilled PPO achieves a better safety--efficiency balance, with smoother progress, fewer unnecessary deviations, and more reliable goal-oriented motion. The BT baseline remains conservative and safe, whereas the distilled learned policy preserves safety while adapting more efficiently to local geometry changes. Together, these results support the claim that BT-guided distillation with latent world-model cues improves practical navigation behavior under dynamic underwater conditions.

\section{Conclusion}
\label{sec:conclusion}

This paper presented an observation-only hybrid autonomy framework for UUV navigation in dynamic underwater environments, combining persistent sonar-based occupancy mapping, clearance-aware global planning, and a learned local controller trained with BT-guided distillation. To improve robustness under partial observability, this work integrated a latent world model and uncertainty-calibrated supervision during policy optimization. Across multi-seed experiments in high-fidelity simulation, the framework improved reliability and safety-efficiency balance relative to non-distilled RL baselines, while maintaining strong performance under increasing dynamic obstacle density and speed. The results demonstrated that structured planning and uncertainty-aware learning can be integrated into a closed-loop autonomy stack without privileged map access. This reduces unsafe exploration, improves stability near obstacle interactions, and yields more consistent navigation than non-distilled RL baselines.
These results are simulation evidence rather than a claim of real-world readiness. The most vulnerable transfer components are sonar artifacts, hydrodynamic mismatch and currents, localization drift, perception noise, and actuator delay/saturation. Future work will validate these components through hardware-in-the-loop, tank, and open-water tests, and will investigate online adaptation and risk-aware replanning for long-horizon robustness under real sensing artifacts.



\bibliographystyle{IEEEtran}
\bibliography{references}

\appendix
\begin{table}[H]
\centering
\caption{Reproducibility parameters.}
\label{tab:app_repro}
\tiny
\setlength{\tabcolsep}{1.8pt}
\renewcommand{\arraystretch}{0.78}
\resizebox{\columnwidth}{!}{%
\begin{tabular}{p{0.27\columnwidth}p{0.27\columnwidth}p{0.21\columnwidth}p{0.37\columnwidth}}
\toprule
Item & Value & Item & Value \\
\midrule
\multicolumn{4}{l}{\textit{Platform and dynamics}}\\
Vehicle & BlueROV-class, 6 thrusters & Control DOF & Surge, sway, heave, yaw-rate \\
Timing & 0.02~s step, 2 substeps & Rigid body & USD mass; no mass randomization \\
Fluid/buoyancy & $g=9.81$, $\rho=997$, $V=0.0113459$~m$^3$ & Drag/coBM & Drag coeff. 0.3; coBM 0.01~m \\
Added mass diag & [5.5, 12.7, 14.57, 0.12, 0.12, 0.12] & Linear damping & [4.03, 6.22, 5.18, 0.07, 0.07, 0.07] \\
Quadratic damping & [18.18, 21.66, 36.99, 1.55, 1.55, 1.55] & Thrusters & $K_F=4.4{\times}10^{-7}$, $K_M=1.3678{\times}10^{-9}$, $\tau=0.01$, max 3900 \\
\midrule
\multicolumn{4}{l}{\textit{Environment, sensing, and safety}}\\
Training setup & 8 envs, 1200 iters, horizon 3000 & Spawn range & $x[2,18]$, $y[1.2,4.8]$, $z[-1.2,-0.4]$ \\
Goal randomization & Enabled; min start-goal 3.0~m & Sensing & Sonar 256 rays/360$^\circ$/12~m; depth 128$\times$72/90$^\circ$/10~m \\
Safety & Goal 1.0~m; collision 0.1~m; TTC 0.18~s; out-of-tank on & Planner map & obstacle inflate 0.08~m; overlay clearance floor 0.1~m \\
\midrule
\multicolumn{4}{l}{\textit{Learning and auxiliary models}}\\
PPO architecture & MLP [256,256] ELU; rollout 64 & PPO optimizer & lr $3{\times}10^{-4}$; 6 epochs; 4 mini-batches; clip 0.2 \\
PPO losses & $\gamma=0.99$, GAE $\lambda=0.95$, entropy 0.003, value 1.5, grad norm 1.0 & BT distillation & batch 1024, 2 grad steps, lr $1.5{\times}10^{-4}$; weight 1.0$\rightarrow$0.2$\rightarrow$0 \\
World model & latent 64, hidden [128,128], target dim 24 & WM optimizer & batch 2048, 2 grad steps, lr $3{\times}10^{-4}$, recon/dyn 1.0/0.2 \\
Uncertainty scaling & EMA 0.05, gain 0.8, max 2.0 & Interpretation & decoder-variance proxy; not calibrated safety probability \\
\bottomrule
\end{tabular}
}
\vspace{-8pt}
\end{table}

\end{document}